\documentclass[10pt,twocolumn,letterpaper]{article}

\usepackage{multirow}
\usepackage{makecell}
\usepackage[normalem]{ulem}
\useunder{\uline}{\ul}{}

\usepackage[pagenumbers]{cvpr} 

\definecolor{cvprblue}{rgb}{0.21,0.49,0.74}
\definecolor{amber}{rgb}{1.0, 0.49, 0.0}
\definecolor{dodgerblue}{RGB}{30, 144, 255}
\definecolor{violet}{RGB}{238,130,238}
\definecolor{my_green}{RGB}{113,173,71}
\definecolor{my_blue}{RGB}{44,115,182}

\newcommand{\reffig}[1]{\textcolor{black}{Figure~\ref{fig:#1}}} 
\newcommand{\refsec}[1]{\textcolor{black}{Section~\ref{sec:#1}}}
\newcommand{\reftab}[1]{\textcolor{black}{Table~\ref{tab:#1}}}

\usepackage[pagebackref,breaklinks,colorlinks,allcolors=cvprblue]{hyperref}
\usepackage{float}
\usepackage{xcolor}
\usepackage{hyperref}
\usepackage{csquotes}
\usepackage{xurl}

\title{DuetSVG: Unified Multimodal SVG Generation with Internal Visual Guidance}

\author{
Peiying Zhang$^{1}$\footnotemark[1] \quad 
Nanxuan Zhao$^{2}$ \quad 
Matthew Fisher$^{2}$ \quad \\
Yiran Xu$^{2}$ \quad  
Jing Liao$^{1}$ \quad  
Difan Liu$^{2}$ \\ \\
$^{1}$City University of Hong Kong \quad
$^{2}$Adobe Research \quad
}

\begin{document}
\begin{sloppypar}

\twocolumn[{
\renewcommand\twocolumn[1][]{#1}
\maketitle
    \vspace{-1.9em}
    \centering
    \includegraphics[width=\textwidth]{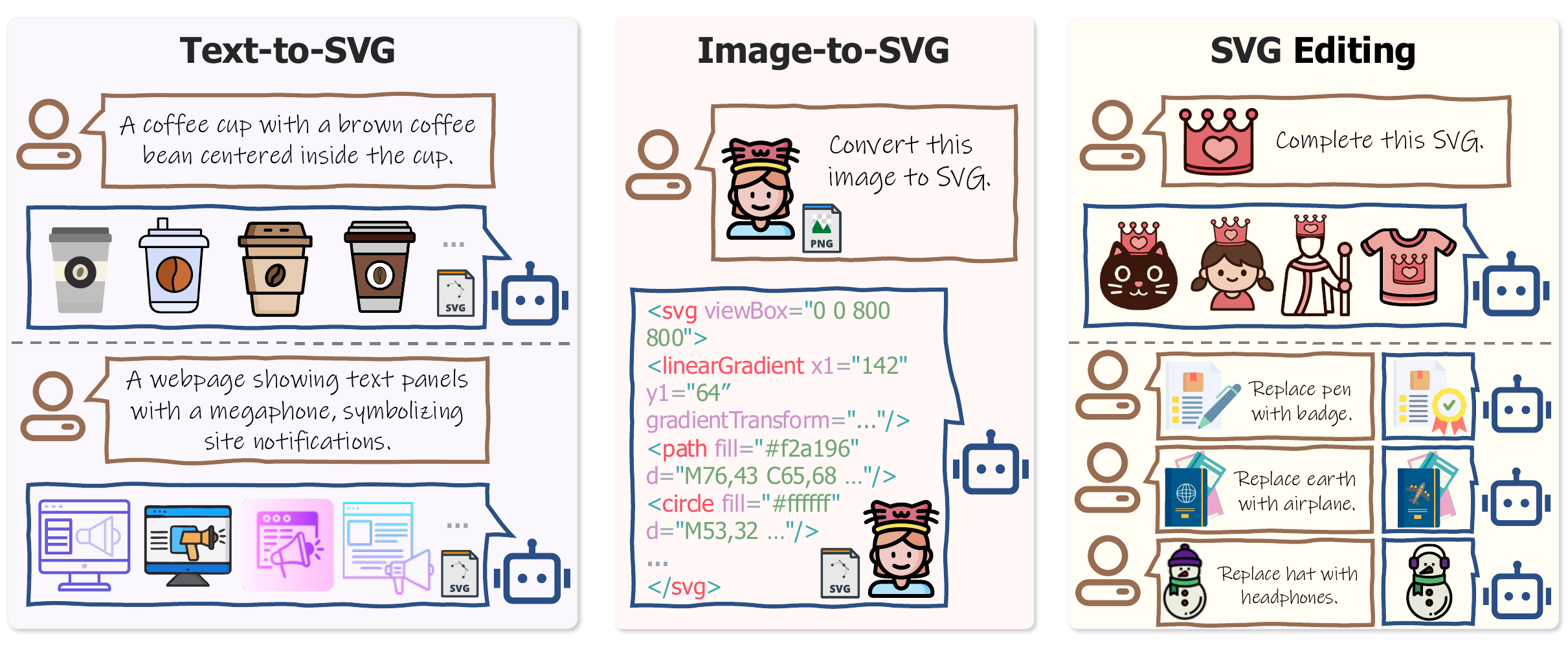}
    \vspace{-1.6em}
    \captionsetup{type=figure}
    \captionof{figure}{
    \textbf{We propose a unified multimodal model, DuetSVG, for SVG generation.}
    DuetSVG acts as a versatile framework across text-to-SVG, image-to-SVG, and SVG editing tasks, demonstrating strong semantic alignment and high-quality SVG generation.
    }
    \vspace{1.2em}
    \label{fig:teaser}
}]

{
  \renewcommand{\thefootnote}%
    {\fnsymbol{footnote}}
  \footnotetext[1]{Work done during internship at Adobe Research.}
}

\begin{abstract}

Recent vision-language model (VLM)-based approaches have achieved impressive results on SVG generation.
However, because they generate only text and lack visual signals during decoding, they often struggle with complex semantics and fail to produce visually appealing or geometrically coherent SVGs.
We introduce DuetSVG, a unified multimodal model that jointly generates image tokens and corresponding SVG tokens in an end-to-end manner. DuetSVG is trained on both image and SVG datasets. At inference, we apply a novel test-time scaling strategy that leverages the model's native visual predictions as guidance to improve SVG decoding quality.
Extensive experiments show that our method outperforms existing methods, producing visually faithful, semantically aligned, and syntactically clean SVGs across a wide range of applications.
The project page is \textcolor{blue}{ \url{https://intchous.github.io/DuetSVG-site}}.

\end{abstract}    
\section{Introduction}
\label{sec:intro}

Scalable Vector Graphics (SVG) are widely used in graphic design, digital art, publishing, and motion graphics. Compared to raster images, SVGs offer resolution-independent rendering, efficient storage, and intuitive editability via control point manipulation. However, creating high-quality vector graphics remains a complex and time-consuming process, even for experienced designers.

With the increasing prominence of large language models (LLMs) and vision-language models (VLMs), recent approaches \cite{rodriguez2023starvector, yang2025omnisvg, xing2025empowering, wang2025internsvg} have leveraged the textual nature of SVGs by formulating SVG generation as a text generation problem and finetuning large text generation models \cite{Qwen2.5-VL, zhu2025internvl3, Qwen3-VL}. These methods produce impressive results on SVG generation tasks such as text-to-SVG.
However, SVGs differ fundamentally from plain text, as they contain an additional dimension—the visual aspect. Formulating SVG generation purely as a text generation task introduces inherent limitations and hampers the performance of existing LLM-based methods.
First, existing approaches are unimodal generative models that produce only text tokens.
Minor errors, such as incorrect predictions of path coordinates, may appear negligible in text space but can lead to catastrophic failures in the rendered SVG.
The absence of visual guidance during text generation represents a main limitation for these models on tasks such as text-to-SVG and SVG completion.
Second, LLM-based SVG generation models exhibit poor generalization beyond their training distribution. As unimodal generative models, they are restricted to training on the relatively small amount of available SVG data and cannot leverage the abundance of high-quality raster image datasets (such as text-image pairs and image editing data), which significantly constrains their generalization capability.

To address the above-mentioned challenges, we propose a novel multimodal generative model that produces native image outputs for SVG generation and editing.
Our model generates a multimodal sequence consisting of image tokens and SVG text tokens. 
The generated image tokens serve as internal visual guidance during SVG token generation, enabling more coherent and visually grounded SVG results. 
Moreover, this multimodal generation framework unlocks new capabilities beyond those of purely text-based vector models. For example, by jointly training on tasks such as text-to-image and text-to-SVG, the model can leverage large-scale text-image datasets for pretraining, which greatly improves generalization and text-SVG alignment.
The multimodal generative nature of our method also simplifies verifier design for test-time scaling. We further introduce a novel and efficient scaling strategy that enhances the model's reliability and robustness during inference.
On the input side, our model accepts multimodal conditions, including raster images, SVG code, and text prompts, and supports a wide range of tasks such as text-to-SVG, image-to-SVG, SVG completion, and SVG editing.

\paragraph{Contributions.}
We present the first unified multimodal generative model for SVG generation. 
Our native visual guidance enables visually grounded SVG generation and allows SVG tasks to be trained on image datasets. 
We further introduce a novel test-time scaling strategy that efficiently improves model reliability.
We demonstrate that DuetSVG achieves much higher quality than state-of-the-art (SoTA) methods on multiple benchmarks.

\section{Related Work}
\label{sec:related_work}

\subsection{Optimization-based SVG Generation}

Classic image vectorization methods \cite{bessmeltsev2019vectorization, tian2022survey, chakraborty2025image, kopf2011depixelizing, selinger2003potrace, favreau2017photo2clipart, hoshyari2018perception, dominici2020polyfit, ma2022towards} rely on fitting algorithms to reconstruct vector graphics from raster images. Although these methods can accurately reproduce the overall appearance of an image, they often generate redundant paths, imprecise control points, and struggle to handle path occlusions.

Recent approaches utilize pre-trained vision-language models (VLMs), such as CLIP \cite{radford2021learning} and diffusion models \cite{rombach2022high}, to directly optimize SVG paths through differentiable rendering \cite{li2020differentiable}.
CLIP-based methods \cite{frans2022clipdraw, schaldenbrand2022styleclipdraw, song2022clipvg, vinker2022clipasso} optimize SVG representations by maximizing image-text alignment within CLIP's latent space.
To leverage the strong visual and semantic priors of text-to-image diffusion models, several methods employ score distillation sampling \cite{poole2022dreamfusion, wang2023prolificdreamer} to optimize static \cite{jain2022vectorfusion,iluz2023word,xing2023diffsketcher,xing2023svgdreamer,zhang2024text} or animated \cite{gal2023breathing, wu2024aniclipart} SVGs to align with textual descriptions. 
However, these methods often require tens of minutes to optimize a single SVG, making them impractical for real-world applications.
More importantly, as they are not trained on vector graphics data, they often produce fragmented paths, inconsistent topology, and redundant control points, which complicate subsequent editing and manipulation.

\subsection{Learning-based SVG Generation}

Early approaches to SVG generative modeling formulated SVGs as sequences of geometric primitives, using VAEs \cite{carlier2020deepsvg, wang2021deepvecfont, wang2023deepvecfont}, or diffusion models \cite{thamizharasan2024vecfusion, das2023chirodiff} as the foundational generative models. 
VecFusion \cite{thamizharasan2024vecfusion} first employs a raster diffusion model to generate an image, followed by a vector diffusion model conditioned on the raster output to produce vector graphics. However, the lack of end-to-end training limits generalization between the raster and vector models, often leading to inaccurate control points and suboptimal geometry.

Recent advances in large language models have inspired SVG generation approaches \cite{wu2023iconshop, tang2024strokenuwa, rodriguez2023starvector, yang2025omnisvg, xing2025empowering, wang2025internsvg, wu2025chat2svg, xing2025reason, rodriguez2025rendering} to represent SVG scripts as discrete text tokens through specialized tokenization schemes, enabling the autoregressive generation of SVG command sequences.
These methods finetune LLMs or VLMs on SVG datasets for tasks such as Text-to-SVG, Image-to-SVG and SVG editing, achieving impressive results.
However, these models exhibit limited generalization because they are trained on relatively small SVG datasets.
Since SVG generation is formulated as a text generation task, they also lack visual guidance during inference, which further constrains their output quality.
Concurrent work RoboSVG \cite{wang2025robosvg} relies on external VLMs to generate additional multimodal conditions as input, which may introduce inconsistencies between models.
In contrast, our unified multimodal generative model co-generates image and SVG tokens within an end-to-end architecture, enabling the use of large-scale text-image data and improving visual grounding during SVG decoding.

\subsection{Unified Multimodal Generation}

Recent unified multimodal models have made substantial progress, showing that a single architecture can both understand and generate multiple modalities, including fully autoregressive \cite{chen2025janus, cui2025emu3} and AR-diffusion fused \cite{deng2025emerging, xie2025show} paradigms. We refer readers to \cite{zhang2025unified} for a more comprehensive review.
However, as SVG scripts differ significantly from natural language in structure, SoTA VLMs still fail to produce high-quality vector graphics without SVG-specialized training.

\section{Method}
\label{sec:method}

\subsection{Task Definition}

An SVG file contains a sequence of paths with drawing commands (e.g., \texttt{M}, \texttt{C}, \texttt{Q}, \texttt{Rect}), numeric coordinates, and style attributes (e.g., \texttt{fill}, \texttt{stroke}).
Previous work \cite{yang2025omnisvg, rodriguez2023starvector} formulate SVG generation as a code generation task and fine-tunes language models to output SVG as text tokens.
However, these methods are limited to SVG-based training data and generalize poorly to complex or out-of-distribution inputs. The absence of visual guidance during SVG generation further constrains their capabilities.
In contrast, we train a unified multimodal generative model that produces both image and SVG tokens. The image modality captures appearance, and the SVG modality learns shape geometry and layer structure.
Formally, given a multimodal conditioning input $\mathbf{x}$, which may include text prompts, images, and SVG code, our model generates a mixed-modality target sequence $\mathbf{z}$ consisting of image tokens $z^{\text{img}}$ and SVG tokens $z^{\text{svg}}$, where $\mathbf{z} = [\langle\texttt{IMG}\rangle, z^{\text{img}}_{1:I}, \langle/\texttt{IMG}\rangle, \langle\texttt{SVG}\rangle, z^{\text{svg}}_{1:S}, \langle/\texttt{SVG}\rangle]$.
Training maximizes the unified next-token objective:
\begin{equation}
  P_{\theta}(\mathbf{z} \mid \mathbf{x}) = \prod_{t=1}^{T} p_{\theta}\left( z_t \mid z_{<t}, \mathbf{x} \right)
\end{equation}
where $\theta$ denotes the model parameters.

\begin{figure}[tbp]
  \centering
  \includegraphics[width=0.9\columnwidth]{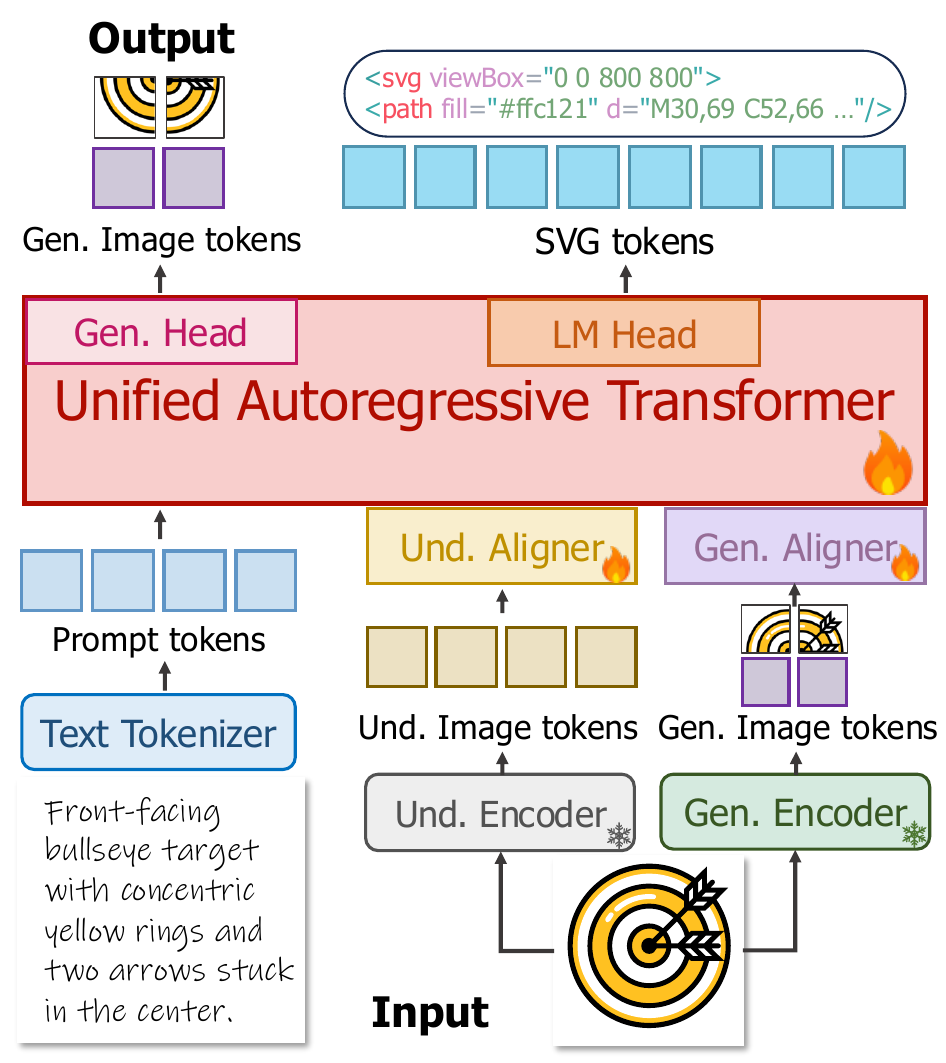}
  \caption{ 
  \textbf{Model architecture of DuetSVG.}
  As a unified model, DuetSVG accepts multimodal inputs, including text prompts, SVG code and raster images.
  We use Janus-Pro text tokenizer ~\cite{chen2025janus} for text prompts.
  For images, an Understanding (Und.) Encoder extracts semantic features, while a Generation (Gen.) Encoder converts images into discrete visual tokens.
  Two MLP aligners project the encoder outputs into the same feature space as the text embeddings. 
  A Generation (Gen.) head predicts image tokens, and a language modeling (LM) head predicts SVG tokens.
  }
  \label{fig:pipeline}
\end{figure}

\subsection{Unified Multimodal SVG Generation Model}

\noindent\textbf{Model Architecture.}
Building on recent unified autoregressive models, our architecture follows Janus-Pro \cite{chen2025janus}, which supports multimodal generation of both image and text tokens, as illustrated in \reffig{pipeline}.
Our model accepts multimodal inputs, including text prompts, SVG code, and images. 
We allow different input configurations for different tasks, such as text prompts for text-to-SVG generation and all three modalities for SVG completion.
Text and SVG inputs are embedded using Janus-Pro's text tokenizer. For image inputs, we employ SigLIP \cite{zhai2023sigmoid} as the understanding encoder to extract semantic features, and a VQ tokenizer \cite{sun2024autoregressive} as the generation encoder to convert images into compact discrete embeddings.
Two separate MLP-based aligners, one for understanding and one for generation, map image embeddings into the LLM feature space.
The concatenated multimodal sequence is input to a unified autoregressive transformer with causal attention, enabling the model to learn cross-modal alignment and next-token prediction over visual and SVG modalities. 
We use a generation head to predict image tokens from a visual codebook and an LM head to predict SVG tokens from a text vocabulary.
During training, the parameters of the understanding and generation image encoders are frozen, while all other parameters are trainable.

\noindent\textbf{Training Stages.} 
Leveraging its multimodal generative design, DuetSVG can be trained on both image and SVG datasets. Training proceeds in two stages: text-to-image pretraining and multi-task supervised fine-tuning (SFT).

Because the Janus-Pro base model has limited ability to generate SVG-style images, we begin with large-scale text-to-image (T2I) pretraining in \textbf{Stage 1}. 
The objective of this stage is to strengthen the model's capacity to produce visually appealing and clean images characterized by clear geometric primitives and flat colors.
We train on a hybrid corpus of real and synthetic T2I data, including: 
(i) rendered SVG images paired with captions from curated SVG datasets, and
(ii) synthetic images generated by FLUX.1 \cite{labs2025flux}, which takes a text prompt and an SVG reference to produce images that match the reference's style.
This pretraining stage enables the model to learn strong semantic and visual priors from large-scale T2I data, providing a robust initialization for subsequent SVG generation tasks.
As shown in our experiments, T2I pretraining helps the model generalize better to complex text prompts and out-of-distribution inputs when generating SVGs.

In \textbf{Stage 2}, we perform SFT across multiple tasks including T2I, T2SVG, and I2SVG under a unified next-token prediction objective with cross-entropy loss over interleaved multimodal outputs.
For SVG-generation tasks, we arrange the target sequence as image tokens followed by SVG tokens so that, during autoregressive decoding, the image tokens can guide the SVG tokens via causal attention.
To strengthen robustness and improve the model's understanding of structural relationships in the I2SVG task, we apply SVG-specific data augmentations. 
In particular, we randomly modify the rotation, translation, scaling, and color attributes of the target SVG, optionally remove a subset of its paths, and render the modified SVG into an image.
We further apply random dropout to the text and image inputs with probability $10\%$, enabling classifier-free guidance \cite{ho2022classifier} during inference.
Multi-task SFT allows the model to share knowledge across modalities and tasks. For example, T2I and I2SVG can enhance T2SVG in different ways, leading to better model quality and stronger generalization. 
In an \textbf{optional Stage 3}, DuetSVG can be further finetuned for downstream applications such as SVG completion, as detailed in Section \ref{sec:applications}.

\begin{figure}[tbp]
  \centering
  \includegraphics[width=1.0\columnwidth]{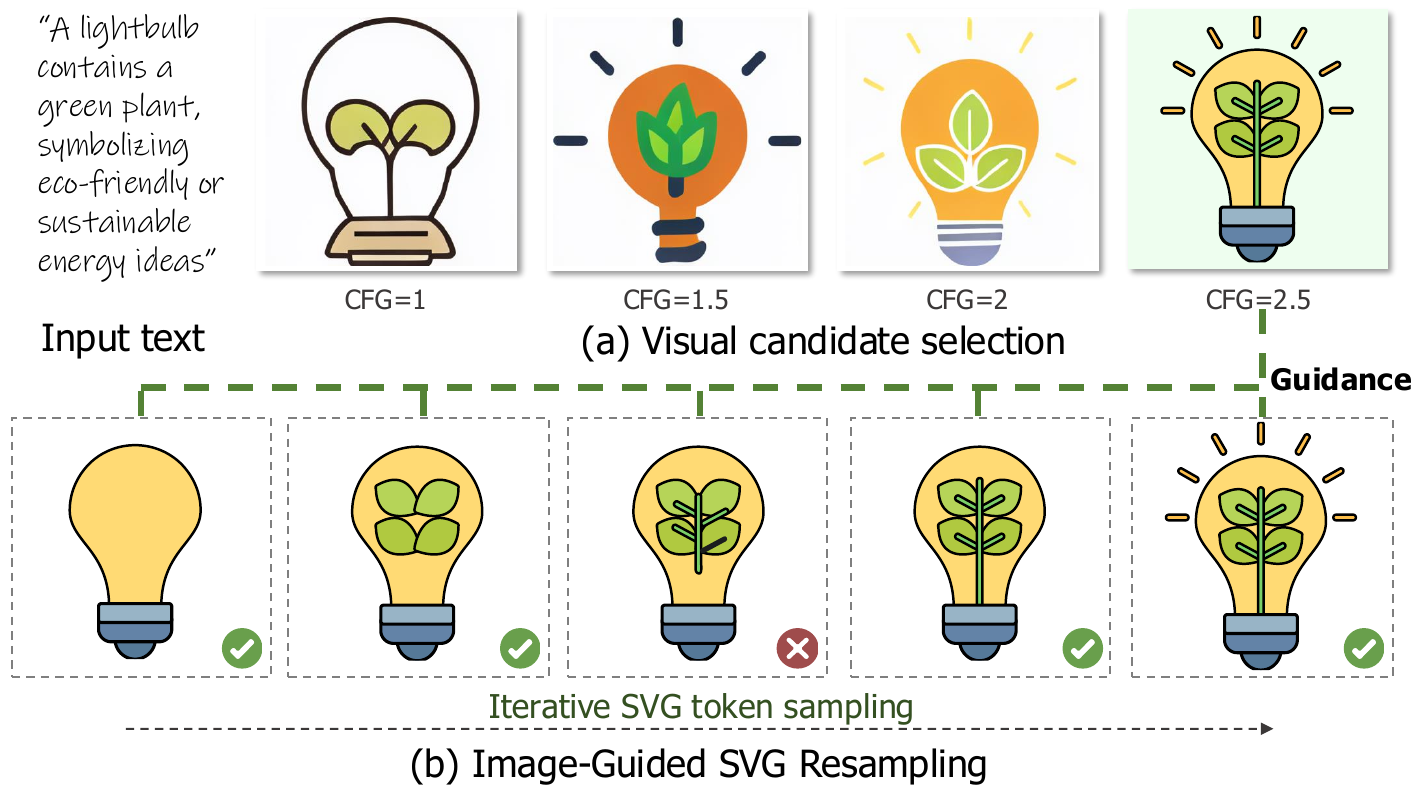}
  \caption{ \label{fig:inference_rs} \textbf{Test-time scaling with image-guided SVG resampling.}
  (a) We first generate $N$ raster image candidates with CFG. Since image-token sequences are much shorter than SVG-token sequences, this step is efficient. A CLIP-based verifier~\cite{radford2021learning} selects the best candidate $I^*$.  
  (b) Using the selected image tokens as internal guidance, we iteratively generate SVG tokens. At each iteration, we render a raster image $R_t$ from the current SVG codes and accept the update only if the LPIPS distance $d(R_t, I^*)$ does not increase; otherwise, we reject and resample.
  }
\end{figure}

\subsection{Test-Time Scaling with SVG Resampling}

For complex SVGs containing thousands of tokens, autoregressive decoding can accumulate sampling errors—such as spurious loops or weakened grounding—often producing suboptimal geometry or even invalid SVGs.
In pure text-based generation model, a common test-time scaling approach is best-of-$N$ sampling: the model produces $N$ complete SVG rollouts, renders each, and a verifier (\eg, CLIP) selects the best result. This strategy is computationally expensive and only reranks after full outputs are generated, providing no visual guidance during decoding.
In contrast, our multimodal model jointly generates image and SVG tokens, enabling a more efficient test-time scaling method with image-guided resampling during inference (see \reffig{inference_rs}).
Our procedure has two stages: (1) selecting the best visual candidate at the image level, and (2) performing image-guided resampling as the SVG generation continues from the stage-1 visual.

\noindent\textbf{Visual Candidate Selection.}
We first generate $N$ visual candidates using classifier-free guidance (CFG):
\begin{equation}
z_t^{\text{CFG}} = z_t^{\text{uncond}} + \gamma \bigl(z_t^{\text{cond}} - z_t^{\text{uncond}}\bigr)
\end{equation}
where $z_t^{\text{cond}}$ and $z_t^{\text{uncond}}$ are the model predictions for image tokens with and without conditioning, respectively, and $\gamma$ is the guidance scale.
Because image-token sequences are much shorter than SVG-token sequences, sampling $N$ candidate images is relatively efficient.
We then score each candidate image using CLIP as the verifier and keep the best one, denoted by $I^\ast$ with corresponding image tokens $z^{\ast}_{\text{img}}$, which can serve as visual guidance during SVG decoding.

\noindent\textbf{Image-Guided SVG Resampling.}
We continue the SVG token generation from the best image tokens $z^{\ast}_{\text{img}}$.
We generate SVG tokens in small chunks with image-guided resampling.
More specifically, at each iteration, we generate $K$ SVG tokens, append them to the current SVG script, and render a provisional raster $R_t$. We then compute its perceptual distance to the best visual candidate $d(R_t, I^\ast)$ using LPIPS \cite{zhang2018unreasonable}. If $d(R_t, I^\ast)$ is less than or equal to $d(R_{t-1}, I^\ast)$, we accept the newly generated tokens; otherwise, we reject them and resample, allowing up to $M$ rejections per SVG.
This image-guided resampling encourages the decoded SVG to remain consistent with the selected visual candidate while avoiding the high cost of best-of-$N$ sampling over long SVG-token sequences. 

By coupling image-level search with chunked, image-guided SVG resampling, our test-time scaling strategy improves semantic alignment and SVG validity at much lower computational cost than naive best-of-$N$ sampling.

\subsection{SVG-Hub Dataset}
\label{sec:svg_dataset}
Existing SVG datasets and benchmarks suffer from limited quality and diversity.
On one hand, many are constructed by vectorizing raster images (e.g., MMSVG \cite{yang2025omnisvg}, InternSVG \cite{wang2025internsvg}), which introduces irregular paths and visual artifacts that compromise SVG structure and regularity.
On the other hand, the accompanying text descriptions are often short and generic, lacking the fine-grained semantics required for high-quality, complex T2SVG generation.

To address these issues, we introduce SVG-Hub-1M, curated from diverse public SVG sources (MMSVG\cite{yang2025omnisvg}, SVGX \cite{xing2025empowering}, and Iconfont\footnote{\url{https://www.iconfont.cn}}) with data cleaning and standardization. 
We remove duplicates, auto-vectorized and blank-rendering SVGs. The SVG-Hub-1M dataset will be released to support future research.
We also conducted experiments on an internal large-scale dataset SVG-Hub-5M. Both datasets consist of high-quality SVGs that are not vectorized results of raster images.

\noindent\textbf{SVG Captioning.} 
Previous SVG datasets typically provide only simple text descriptions, which are insufficient for training models to understand complex semantics and generate semantically aligned SVGs. 
To support T2SVG from semantically rich prompts, we rasterize each SVG and use open-source VLMs (InternVL3 \cite{zhu2025internvl3} and Qwen2.5-VL \cite{Qwen2.5-VL}) to produce captions at three levels of detail: (1) short prompts that capture core semantics, (2) medium descriptions that enumerate semantic elements along with their layout and style, and (3) detailed annotations covering fine-grained shapes, strokes, and colors. 
Our dataset pairs each SVG with a comprehensive set of captions. Additional details of the captioning pipeline are provided in \refsec{captioning_pipeline}.

\noindent\textbf{SVG Tokenization.} 
An SVG file contains geometric primitives and paths with drawing commands and attributes.
We construct a compact and regular representation by (i) removing redundant or invisible elements, (ii) normalizing the canvas to an 800$\times$800 \texttt{viewBox}, and (iii) restricting the command vocabulary to \{\texttt{M}, \texttt{L}, \texttt{C}, \texttt{Q}, \texttt{A}, \texttt{Z}, \texttt{Ellipse}, \texttt{Circle}, \texttt{Polygon}, \texttt{Rect}\}.
We then quantize all coordinates $(x,y)$ and serialize the normalized SVG into a sequence of discrete tokens that includes command, attribute, and quantized coordinate tokens.
Gradient definitions (\texttt{<defs>}) and group-level transformations (\texttt{<g>}) are retained to preserve expressiveness.
These steps standardize the SVG script structure and reduce file size while remaining lossless with respect to rendering.

\begin{table*}[tbp]
  \caption{ \textbf{Quantitative comparison with existing methods on the SVG-Hub-5M test set.} \textbf{Bold} scores indicate the best results among all methods, and \underline{underlined} scores indicate the best results among VLM-based methods without test-time scaling (w/o TTS). 
  }
  \resizebox{\linewidth}{!}{
    \begin{tabular}{cl|cccc|cccccc}
    \Xhline{1.2pt}

    \multicolumn{2}{c|}{\multirow[c]{3}{*}{Method}}                               & \multicolumn{4}{c|}{\textbf{Text-to-SVG}}                                                                                       & \multicolumn{6}{c}{\textbf{Image-to-SVG}}                                                                                                                                                                                 \\ \cline{3-12} 
    \multicolumn{2}{c|}{}                                                       & FID $\downarrow$ & CLIP $\uparrow$ & Aesthetic $\uparrow$ & \begin{tabular}[c]{@{}c@{}}Path\\ Semantics \end{tabular} $\big\uparrow$ & DINO $\uparrow$ & SSIM $\uparrow$ & LPIPS $\downarrow$ & PSNR $\uparrow$ & \begin{tabular}[c]{@{}c@{}}SVG Code\\ Similarity \end{tabular} $\big\uparrow$ & \begin{tabular}[c]{@{}c@{}}Path\\ Semantics \end{tabular} $\big\uparrow$ \\ \hline
    \multicolumn{1}{c|}{\multirow{5}{*}{Optimization}} & VTracer                & -                & -               & -                    & -                                                                   & \textbf{0.968}  & \textbf{0.936}  & \textbf{0.071}     & \textbf{23.623} & 0.788                                                                    & 0.982                                                               \\
    \multicolumn{1}{c|}{}                              & FLUX.1-dev + VTracer   & 46.990           & 25.326          & 5.52                 & 1.216                                                               & -               & -               & -                  & -               & -                                                                        & -                                                                   \\
    \multicolumn{1}{c|}{}                              & Vectorfusion           & 59.354           & 24.519          & 5.05                 & 1.405                                                               & -               & -               & -                  & -               & -                                                                        & -                                                                   \\
    \multicolumn{1}{c|}{}                              & SVGDreamer             & 56.743           & 24.871          & 5.34                 & 1.468                                                               & -               & -               & -                  & -               & -                                                                        & -                                                                   \\
    \multicolumn{1}{c|}{}                              & T2V-NPR                & 54.420           & 25.024          & 5.38                 & 1.924                                                               & -               & -               & -                  & -               & -                                                                        & -                                                                   \\ \hline
    \multicolumn{1}{c|}{\multirow{11}{*}{VLM}}         & GPT-5-Thinking         & 50.122           & 24.950          & 5.39                 & 2.295                                                               & 0.904           & 0.806           & 0.212              & 11.485          & 0.916                                                                    & 2.496                                                               \\
    \multicolumn{1}{c|}{}                              & Gemini-3-Pro           & 48.765           & 25.146          & 5.46                 & 2.412                                                                   & 0.921           & 0.880           & 0.116              & 13.858          & 0.908                                                                        & 2.516                                                                   \\
    \multicolumn{1}{c|}{}                              & Gemini-2.5-Pro         & 57.597           & 24.572          & 5.20                 & 2.154                                                               & 0.885           & 0.697           & 0.275              & 10.237          & 0.887                                                                    & 2.028                                                               \\

    \multicolumn{1}{c|}{}                              & StarVector-8B (w/o FT) & -                & -               & -                    & -                                                                   & 0.691           & 0.489           & 0.388              & 8.920           & 0.862                                                                    & 1.246                                                               \\
    \multicolumn{1}{c|}{}                              & StarVector-8B (FT)     & -                & -               & -                    & -                                                                   & 0.896           & 0.813           & 0.186              & 18.646          & 0.932                                                                    & 1.830                                                               \\
    \multicolumn{1}{c|}{}                              & LLM4SVG-7B (FT)        & 49.318           & 23.304          & 5.24                 & 2.315                                                               & 0.938           & 0.882           & 0.099              & 19.843          & 0.944                                                                    & 2.245                                                               \\
    \multicolumn{1}{c|}{}                              & OmniSVG-3B (w/o FT)    & 93.172           & 21.225          & 4.32                 & 1.780                                                               & 0.812           & 0.701           & 0.294              & 11.577          & 0.893                                                                    & 1.505                                                               \\
    \multicolumn{1}{c|}{}                              & OmniSVG-3B (FT)        & 58.237           & 22.726          & 5.17                 & 2.286                                                               & 0.933           & 0.854           & 0.105              & 19.256          & 0.932                                                                    & 2.164                                                               \\
    \multicolumn{1}{c|}{}                              & Qwen3-VL-8B (FT)       & 43.720           & 23.935          & 5.35                 & 2.525                                                               & 0.947           & 0.910           & 0.090              & 20.915          & 0.950                                                                    & 2.492                                                               \\
    \multicolumn{1}{c|}{}                              & Ours-7B (w/o TTS)      & {\ul 35.066}     & {\ul 25.584}    & {\ul 5.52}           & {\ul 2.772}                                                         & {\ul 0.955}     & {\ul 0.920}     & {\ul 0.082}        & {\ul 22.024}    & {\ul 0.959}                                                              & {\ul 2.558}                                                         \\
    \multicolumn{1}{c|}{}                              & Ours-7B (TTS)          & \textbf{33.574}  & \textbf{26.106} & \textbf{5.56}        & \textbf{2.910}                                                      & 0.962           & 0.928           & 0.075              & 23.590          & \textbf{0.964}                                                           & \textbf{2.724}                                                      \\

    \Xhline{1.2pt}
    \end{tabular}
  }
  \label{tab:table_quality_eval_svgstock}
\end{table*}

\begin{table}[tbp]
  \caption{ \textbf{Quantitative comparison with VLM-based baselines on the SArena-Icon benchmark~\cite{wang2025internsvg}. }
  }
  \resizebox{\linewidth}{!}{
    \begin{tabular}{l|ccc|cccc}
    \Xhline{1.2pt}

    \multicolumn{1}{c|}{\multirow{2}{*}{Method}} & \multicolumn{3}{c|}{\textbf{Text-to-SVG}}               & \multicolumn{4}{c}{\textbf{Image-to-SVG}}                                \\ \cline{2-8} 
    \multicolumn{1}{c|}{}                        & FID $\downarrow$ & FID-C $\downarrow$ & CLIP $\uparrow$ & DINO $\uparrow$ & SSIM $\uparrow$ & LPIPS $\downarrow$ & PSNR $\uparrow$ \\ \hline
    GPT-5-Thinking                               & 14.892           & 4.993              & 25.125          & 0.916           & 0.815           & 0.134              & 11.512          \\
    Gemini-3-Pro                                 & 14.324           & 4.847              & 25.382          & 0.938           & 0.874           & 0.125              & 14.366          \\
    Gemini-2.5-Pro                               & 15.484           & 5.016              & 24.910          & 0.895           & 0.690           & 0.261              & 10.878          \\
    StarVector-8B (w/o FT)                       & -                & -                  & -               & 0.871           & 0.623           & 0.206              & 13.595          \\
    StarVector-8B (FT)                           & -                & -                  & -               & 0.920           & 0.844           & 0.118              & 19.076          \\
    LLM4SVG-7B (FT)                              & 18.815           & 6.784              & 23.138          & 0.946           & 0.910           & 0.092              & 22.160          \\
    OmniSVG-3B (w/o FT)                          & 28.292           & 11.318             & 21.679          & 0.894           & 0.756           & 0.186              & 12.669          \\
    OmniSVG-3B (FT)                              & 19.941           & 7.477              & 22.450          & 0.937           & 0.908           & 0.099              & 21.984          \\
    Qwen3-VL-8B (FT)                             & 16.780           & 5.192              & 23.704          & 0.955           & 0.921           & 0.081              & 22.750          \\
    Ours-7B (w/o TTS)                            & {\ul 12.105}     & {\ul 4.205}        & {\ul 25.416}    & {\ul 0.969}     & {\ul 0.929}     & {\ul 0.069}        & {\ul 23.482}    \\
    Ours-7B (TTS)                                & \textbf{11.712}  & \textbf{3.928}     & \textbf{25.734} & \textbf{0.972}  & \textbf{0.938}  & \textbf{0.060}     & \textbf{24.016} \\ 
    
    \Xhline{1.2pt}
    \end{tabular}
  }
  \label{tab:table_quality_eval_sarena}
\end{table}

\section{Experiments}
\label{sec:experiments}

\subsection{Implementation Details}
We initialize DuetSVG from Janus-Pro-7B~\cite{chen2025janus}.
We resize each image to $384 \times 384$, and then use the generation encoder to encode it into visual tokens with a sequence length of 576, with a codebook of size 16{,}384. 
Each SVG is tokenized and truncated to a maximum of 12{,}000 text tokens.
During the T2I pre-training stage, we use a mixture of real and synthetic T2I data, and train for 80K steps with a batch size of 512.
In the multi-task SFT stage, we jointly train on T2I, T2SVG, and I2SVG data, sampling them with a ratio of 1:5:4, respectively. 
We train this stage for 300K steps with a batch size of 128.
The full training process takes about 14 days on 64 NVIDIA-A100 GPUs.
We use AdamW~\cite{loshchilov2018decoupled} optimizer with $\beta_1 = 0.9$, $\beta_2 = 0.95$ with a learning rate of $1 \times 10^{-5}$ in all stages. 
For the test-time scaling strategy, we set $N=3$ and $M=3$ in our experiments.

\subsection{Experiment Setup}

\noindent\textbf{Dataset and Benchmark.}
We evaluate our method on two benchmarks. 
One is the test split from SVG-Hub-5M which has 9{,}000 samples.
The other one is the SArena-Icon Benchmark~\cite{wang2025internsvg} which contains 6{,}000 samples.

\noindent\textbf{Evaluation Metrics.}
We evaluate the quality of our results from both vector-level and image-level perspectives.
For \textbf{vector-level} evaluation, we measure \textit{Path Semantics} \cite{zhang2024text} by randomly removing 30\% of the SVG paths and computing the drop in CLIP score~\cite{radford2021learning} between the original and modified renderings. A smaller drop indicates that the generated paths are redundant or carry limited semantic meaning.
In I2SVG, we measure \textit{SVG code similarity} by encoding the generated and ground-truth SVG code with Qwen3-Embedding-8B \cite{qwen3embedding} and computing the cosine similarity between their embeddings, which reflects the syntactic quality of the generated SVGs.
For \textbf{image-level} evaluation, we assess the visual fidelity and quality of T2SVG using \textit{FID}~\cite{heusel2017gans}, \textit{FID-CLIP}~\cite{wu2023iconshop}, \textit{CLIP} score \cite{radford2021learning} and the \textit{Aesthetic} score \cite{schuhmann2021improved}. 
For I2SVG, we measure the visual similarity between the rendered SVGs and the input images using \textit{DINO}~\cite{oquab2023dinov2}, \textit{SSIM}~\cite{wang2004image}, \textit{PSNR}, and \textit{LPIPS}~\cite{zhang2018unreasonable}.

\noindent\textbf{Baselines.}
We compare our DuetSVG against a diverse set of baselines, including both optimization-based and learning-based SVG generation methods.
For optimization-based methods, in the T2SVG task, one baseline uses a raster-then-vectorize pipeline: images are generated by FLUX.1-dev~\cite{labs2025flux} and then vectorized by VTracer \cite{pun_vtracer_2025, selinger2003potrace}.
We also evaluate three text-guided SVG optimization methods, VectorFusion \cite{jain2022vectorfusion}, SVGDreamer \cite{xing2023svgdreamer}, and T2I-NPR \cite{zhang2024text}, each optimized with 64 paths. For the I2SVG task, we use VTracer as a baseline.
Learning-based methods include SoTA proprietary and open-source VLMs.
For proprietary VLMs, we use GPT-5-Thinking \cite{achiam2023gpt}, Gemini-3-Pro \cite{deepmind2025gemini3pro} and Gemini-2.5-Pro~\cite{comanici2025gemini}.
For open-source SVG-specific VLMs, we compare with StarVector~\cite{rodriguez2025starvector}, LLM4SVG~\cite{xing2025empowering}, and OmniSVG~\cite{yang2025omnisvg}, and we also include the recently released VLM Qwen3-VL-8B \cite{Qwen3-VL}.
Since these models adopt different backbones and training data, we further \textbf{fine-tune} all open-source VLM baselines on our SVG-Hub-5M dataset to ensure a fair comparison.

\subsection{Comparisons}

\begin{figure*}[tbp]
  \centering
  \includegraphics[width=1.0\linewidth]{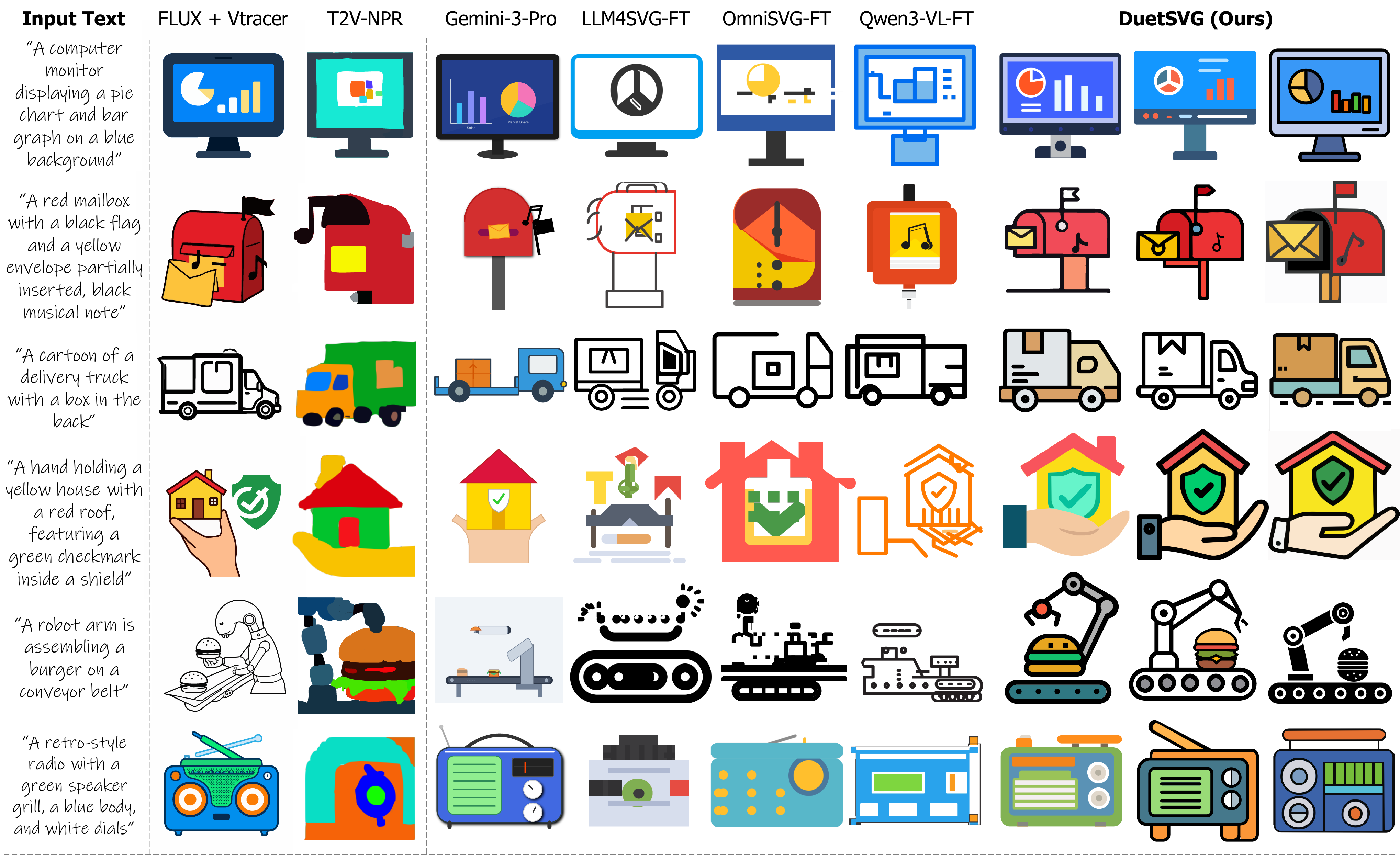}
  \caption{ \label{fig:t2svg_result} \textbf{Qualitative comparison on text-to-SVG generation task.} 
  Our DuetSVG aligns better with the text prompts and generates high-quality visual outputs with detailed structures.
  }
\end{figure*}

\begin{figure*}[tbp]
  \centering
  \includegraphics[width=1.0\linewidth]{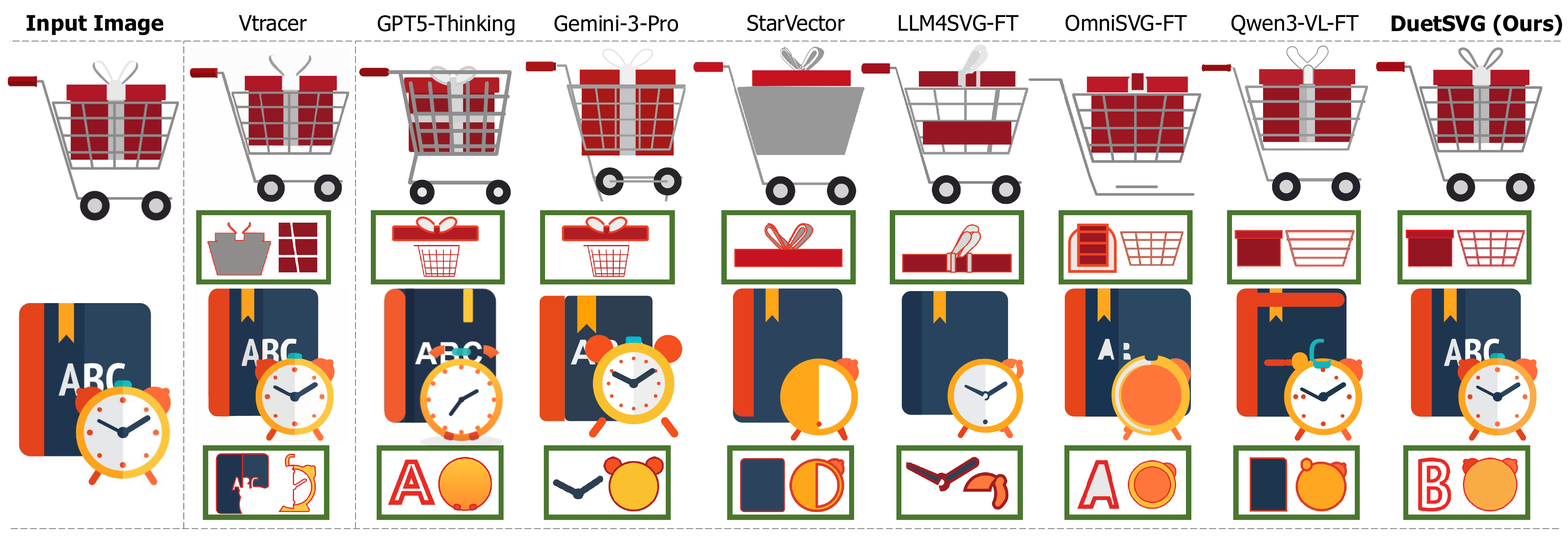}
  \caption{ \label{fig:i2svg_result} \textbf{Qualitative comparison on image-to-SVG conversion task.} Ours preserves more visual details than other previous baselines.}
  \vspace{-3pt}
\end{figure*}

We evaluate the performance of DuetSVG against all baselines qualitatively and quantitatively on both T2SVG and I2SVG tasks.
We report quantitative results on the SVG-Hub-5M test set in \reftab{table_quality_eval_svgstock}, and on the SArena-Icon benchmark in \reftab{table_quality_eval_sarena}.
We also show qualitative comparisons in \reffig{t2svg_result} for T2SVG and \reffig{i2svg_result} for I2SVG.
Overall, DuetSVG achieves consistently stronger performance than existing methods.
We refer readers to \refsec{additional_comparisons_results} for additional comparisons and results.

\noindent\textbf{Text-to-SVG Task.}
As shown in \reftab{table_quality_eval_svgstock} and \reftab{table_quality_eval_sarena}, DuetSVG consistently outperforms all T2SVG baselines across all metrics.

Optimization-based methods are time-consuming and often yield SVGs with redundant paths, artifacts, and disorganized layers.
Regarding learning-based methods, SoTA proprietary VLMs exhibit strong semantic understanding, but they mainly produce combinations of oversimplified primitives (\eg, circles and rectangles) and struggle with spatial layout and fine-grained geometric details (see ~\reffig{t2svg_result}). It indicates that precise SVG generation remains challenging for general VLMs.
For open-source VLMs, evaluating their public checkpoints reveals severe overfitting: they often fail to produce valid SVGs for inputs outside their training distribution. To enable a fair comparison, we fine-tune all VLMs on our SVG-Hub-5M training set. As shown by the ``w/o FT'' and ``FT'' in \reftab{table_quality_eval_svgstock} and \reftab{table_quality_eval_sarena}.
However, these fine-tuned VLMs still struggle to produce semantically accurate SVGs with the complex geometric detail, as shown in \reffig{t2svg_result}.
A major reason is their text-centric design: SVGs are treated purely as text sequences, so training mainly enforces syntactic correctness of the SVG code while providing no supervision on the visual appearance.

In contrast, DuetSVG produces diverse, visually appealing SVGs with clear structure and semantically rich content.
We further analyze the model's generalization in \refsec{generaliz_eval}.

\noindent\textbf{Image-to-SVG Task.}
Classical image vectorization techniques achieve strong reconstruction scores by densely fitting paths to the image, but they generate verbose and inefficient code (reflected by low \textit{SVG Code Similarity} in \reftab{table_quality_eval_svgstock}), lack layer organization (low \textit{Path Semantics} in \reftab{table_quality_eval_svgstock}), and struggle with overly complex vector elements (shown in the green boxes in \reffig{i2svg_result}).
VLM-based baselines struggle to accurately reconstruct complex shapes and fine details, often resulting in inaccurate spatial layouts and simplified fine-grained geometry.
In contrast, DuetSVG produces visually faithful and syntactically clean SVGs, as shown in \reffig{i2svg_result}.

\begin{table}[tbp]
  \caption{ \textbf{Ablation study.}
  (a) Removing visual output brings an notable degradation for all metrics, indicating the effectiveness of our unified multimodal generation design.
  (b) Without T2I pretraining, we observe a clear drop in T2SVG quality, showing that T2I alignment learned during pretraining is crucial.
  (c) Test-time scaling further boosts the quality in both T2SVG and I2SVG tasks.
  }
  \resizebox{\linewidth}{!}{
    \begin{tabular}{l|cc|ccc}
    \Xhline{1.2pt}

    \multicolumn{1}{c|}{\multirow{2}{*}{Method}} & \multicolumn{2}{c|}{\textbf{Text-to-SVG}} & \multicolumn{3}{c}{\textbf{Image-to-SVG}}              \\ \cline{2-6} 
    \multicolumn{1}{c|}{}                        & FID $\downarrow$     & CLIP $\uparrow$    & DINO $\uparrow$ & SSIM $\uparrow$ & LPIPS $\downarrow$ \\ \hline
    w/o Internal Visual Guidance                            & 51.482               & 23.260             & 0.939           & 0.878           & 0.096              \\
    w/o T2I Pretraining                          & 36.953               & 25.122             & -               & -               & -                  \\
    w/o Test-time Scaling                      & 35.066               & 25.584             & 0.955           & 0.920           & 0.082              \\
    Ours (Full)                               & \textbf{33.574}      & \textbf{26.106}    & \textbf{0.962}  & \textbf{0.928}  & \textbf{0.075}     \\ 

    \Xhline{1.2pt}
    \end{tabular}
  }
  \label{tab:ablation_study}
\end{table}

\begin{figure}[tbp]
  \centering
  \includegraphics[width=\columnwidth]{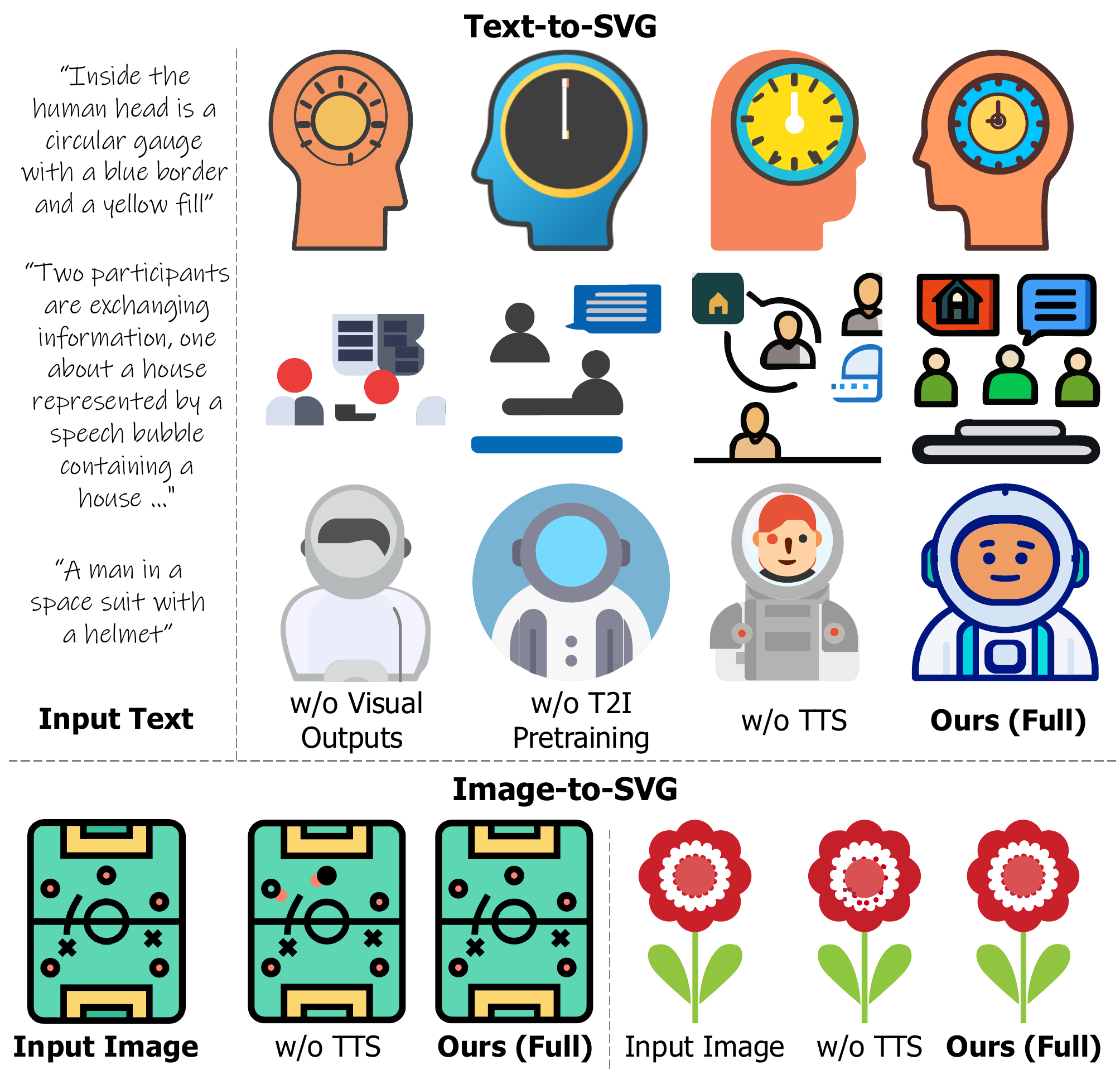}
  \caption{ \label{fig:ablations} \textbf{Ablation results.} 
  (a) Without \textbf{visual output}, SVG-only generation struggles to produce semantically accurate and structurally coherent SVGs. 
  (b) Without \textbf{T2I pretraining}, the model lacks strong visual and semantic priors, resulting in simpler and less visually appealing SVGs.
  (c) Without \textbf{test-time scaling (TTS)}, autoregressive decoding may introduce local geometric distortions when handling complex inputs.
  }
\end{figure}

\begin{figure}[tbp]
  \centering
  \includegraphics[width=\columnwidth]{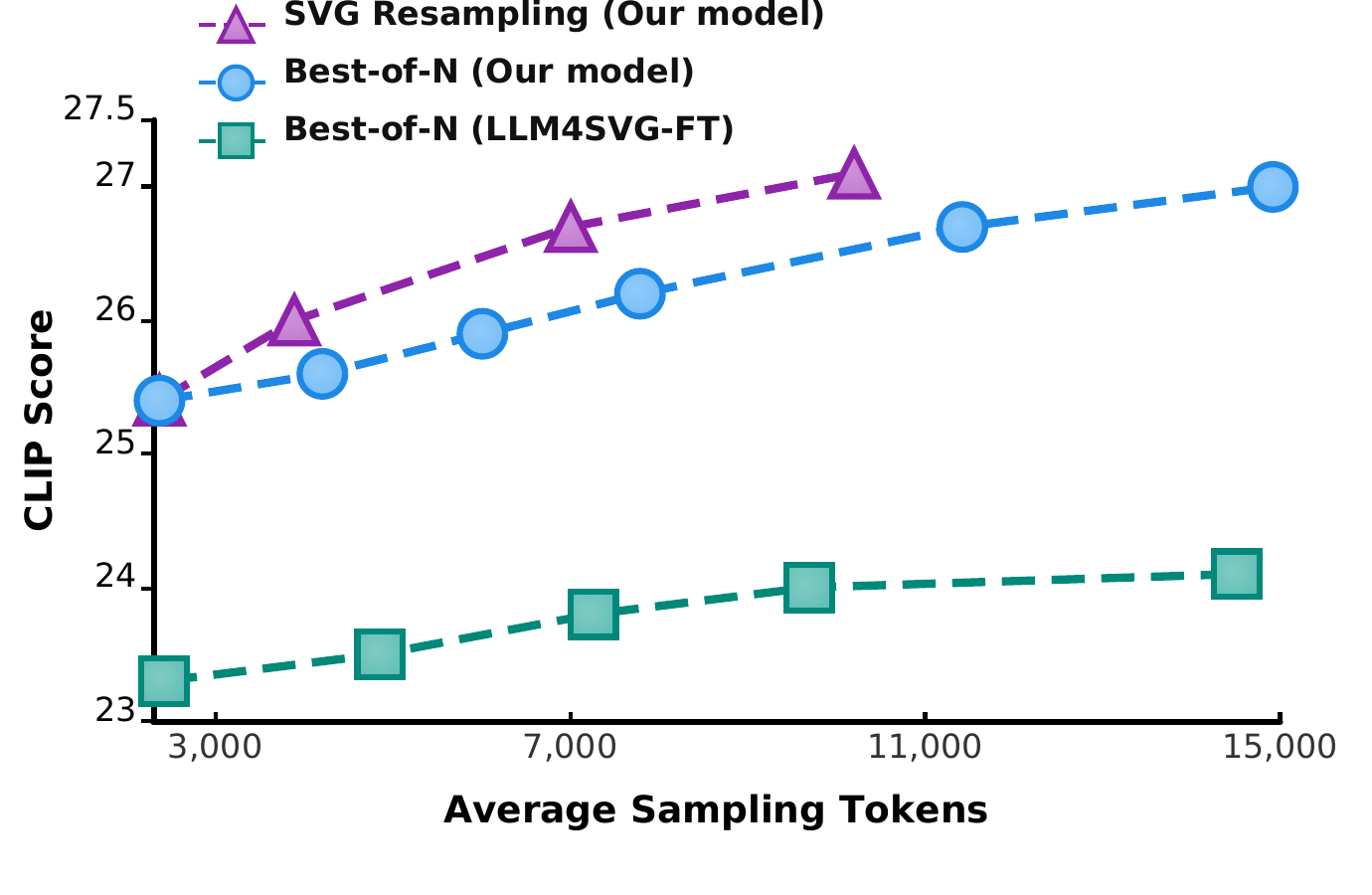}
  \caption{ \label{fig:test-time_abl} \textbf{Sampling efficiency comparison of test-time scaling strategies.}
  Our strategy leverages image-level candidate selection and SVG resampling to achieve efficient test-time scaling, with substantially lower sampling cost than best-of-$N$.
  }
\end{figure}

\subsection{Ablation Study}
\label{sec:ablation_study}

\noindent\textbf{Benefits of Internal Visual Guidance.}
To assess the effect of multimodal SVG generation, we train a variant that shares the same training configuration as DuetSVG but only decodes SVG tokens, without generating image tokens.
As shown in \reffig{ablations}, this SVG-only baseline exhibits failure modes similar to prior text-centric VLMs, struggling to produce semantically accurate and structurally coherent SVGs.
We further observe that it performs worse than the fine-tuned Qwen3-VL-8B in \reftab{ablation_study}, indicating that our backbone is weaker as a pure language model than Qwen3-VL-8B.
Yet DuetSVG still outperforms fine-tuned Qwen3-VL-8B when equipped with image modality, showing that visual guidance is crucial for high-quality SVG generation.

\noindent\textbf{Ablation of T2I Pretraining.}
To validate the effectiveness of the T2I pre-training stage, we conduct an ablation in which this stage is removed.
In~\reftab{ablation_study}, we observe that T2I pre-training helps the model acquire stronger visual and semantic priors, enabling it to generate more visually appealing SVG-style images and more complex SVGs.

\noindent\textbf{Ablation of TTS with SVG Resampling.}
In this ablation study, we evaluate DuetSVG with our test-time scaling (TTS) strategy, DuetSVG with best-of-$N$ TTS, and LLM4SVG with best-of-$N$ TTS.
The \textit{CLIP vs Compute} graph is shown in \reffig{test-time_abl}. Compared to best-of-$N$, our novel two-stage TTS with image-guided SVG resampling is much more efficient. Text-centric VLMs still perform much worse with TTS enabled.

\begin{figure}[tbp]
  \centering
  \includegraphics[width=\columnwidth]{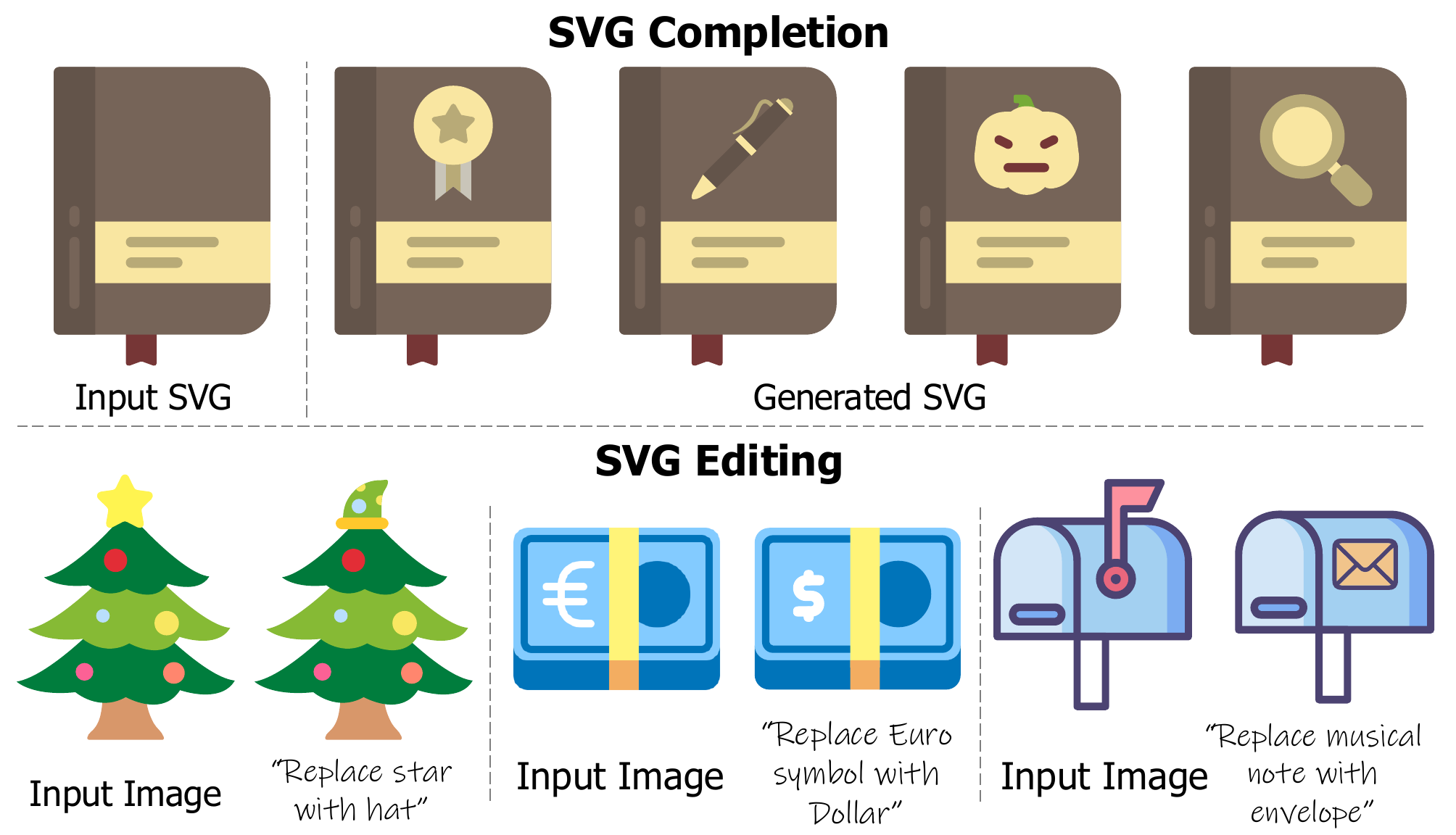}
  \caption{ \label{fig:application} 
  \textbf{Downstream applications.}
  After fine-tuning, DuetSVG supports:
  (a) \textit{SVG completion}: given the partial SVG and its rendered image, DuetSVG completes coherent and visually appealing SVGs.
  (b) \textit{SVG editing}: DuetSVG enables instruction-based editing for SVGs.
  }
\end{figure}

\section{Applications}
\label{sec:applications}
To support downstream applications, DuetSVG can be further fine-tuned for SVG completion and semantic SVG editing.
For \textbf{SVG completion}, we randomly mask a subset of paths in an SVG and provide the model with both the partial SVG and its rendered image as inputs. The model is trained to infer the complete image and the full SVG script. As shown in \reffig{application}, DuetSVG can effectively complete coherent and visually appealing SVGs.
For semantic \textbf{SVG editing}, since no large-scale paired SVG editing dataset is available, we construct a synthetic dataset using a powerful image editing model.
Concretely, for each SVG rendering $I_a$ from our dataset and a text instruction, we use Gemini-2.5-Flash~\cite{comanici2025gemini} to produce an edited image $I_e$ that reflects the instruction.
During training, we take $I_e$ and an inverse editing instruction as inputs, and train DuetSVG to reconstruct the original image $I_a$ and its corresponding SVG script.
As shown in \reffig{application}, DuetSVG can perform high-level semantic SVG editing, modifying SVG content according to the text instruction.

\section{Conclusion}
\label{sec:conclusion}

In this paper, we presented DuetSVG, a unified multimodal model that generates both image tokens and SVG tokens rather than treating SVGs as pure text.
By combining large-scale T2I pre-training with multi-task SFT on T2I, T2SVG, and I2SVG, DuetSVG leverages rich visual priors to achieve stronger text-SVG alignment and to produce high-quality SVGs.
We further introduced a vision-aware test-time scaling strategy that uses the internal visual predictions to guide SVG decoding, improving robustness and reliability.
DuetSVG supports a wide range of SVG generation and editing tasks. 
With advanced VLM techniques, our framework could further accelerate generation, and the proposed training and inference strategies can be used to train larger-scale unified multimodal models (\eg, Emu3.5 \cite{cui2025emu3}), which we leave for future work.

{
    \small
    \bibliographystyle{ieeenat_fullname}
    \bibliography{main}
}

\appendix
\clearpage
\setcounter{page}{1}
\maketitlesupplementary

\section{Overview}

In this supplementary material, we provide additional details and evaluations, including:
\begin{itemize}[leftmargin=*]
  \item Generalization evaluation (\refsec{generaliz_eval}).
  \item Additional comparisons and results (\refsec{additional_comparisons_results}).
  \item Limitation (\refsec{limitation}).
  \item Details of the captioning pipeline (\refsec{captioning_pipeline}).
\end{itemize}

\section{Additional Evaluation}

\subsection{Generalization Evaluation}
\label{sec:generaliz_eval}

We evaluate our model's generalization capability using \textit{Novelty} and \textit{Uniqueness} scores following IconShop \cite{wu2023iconshop}.
Two SVGs are considered identical if the cosine similarity between their CLIP image embeddings is at least 0.98. 
Under this criterion, \textit{Uniqueness} is the fraction of generated samples that appear exactly once within the generated set. 
\textit{Novelty} is the fraction of generated samples that have no match in the \textit{SVG-Hub-5M} training corpus: for each generated SVG, we render it to an image, retrieve its nearest neighbor in \textit{SVG-Hub-5M} using CLIP cosine similarity, and label the sample as novel if the maximum similarity is below 0.98.

\begin{figure}[tbp]
  \centering
  \includegraphics[width=1.0\columnwidth]{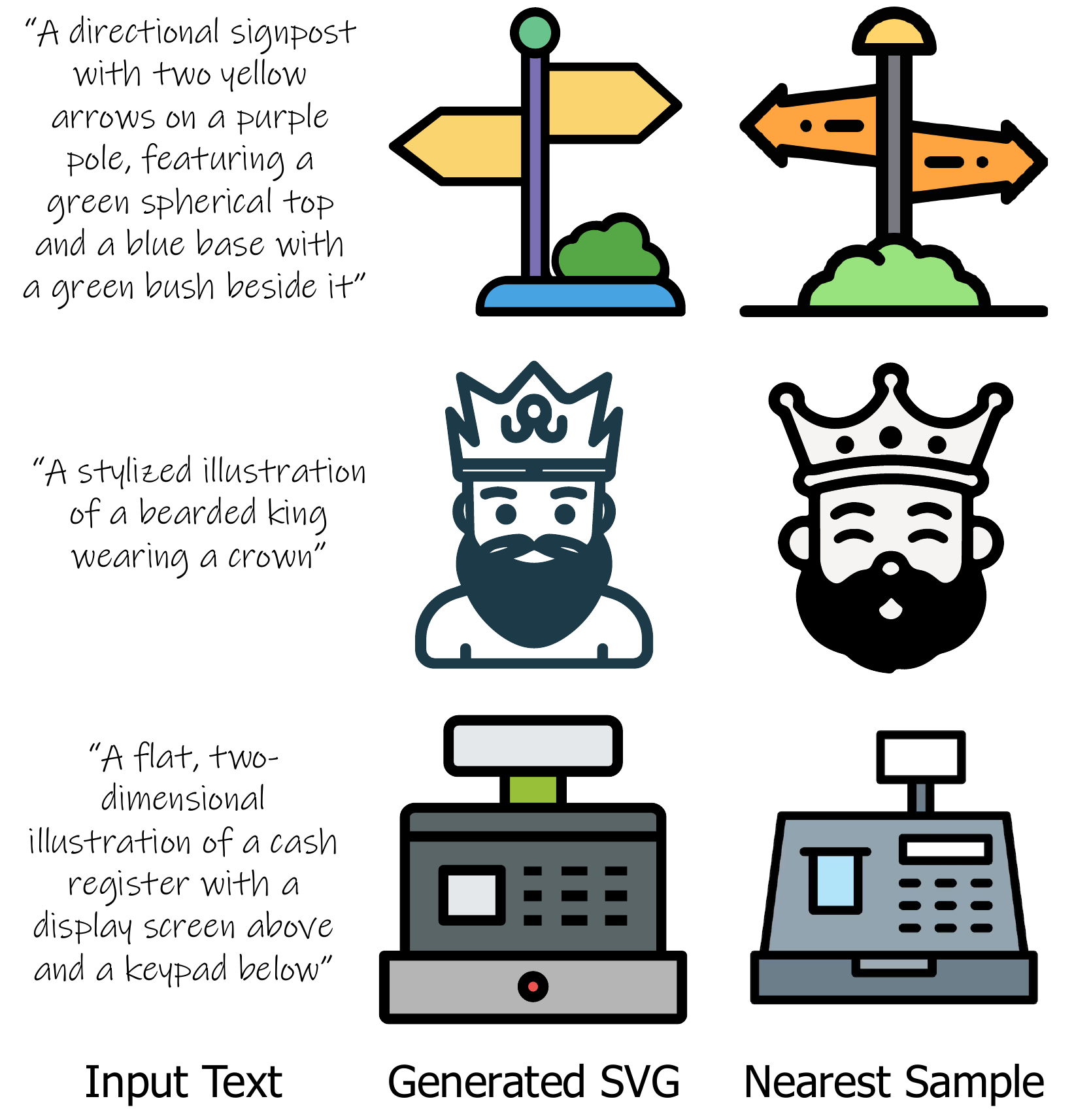}
  \caption{Generated SVGs and their nearest neighbor samples.}
  \label{fig:nearest_sample}
\end{figure}

We randomly select 500 text prompts and generate three SVGs per prompt with different seeds, yielding 1{,}500 samples. 
On this set, our model attains 99.5\% \textit{Novelty} and 99.8\% \textit{Uniqueness}. 
\reffig{nearest_sample} shows generated SVGs alongside their nearest neighbors from \textit{SVG-Hub-5M}, illustrating our model's ability to produce novel and diverse results.

\subsection{Additional Comparisons and Results}
\label{sec:additional_comparisons_results}

\paragraph{Additional Results on SVG-Hub-1M.}
We provide additional quantitative results on the \textit{SVG-Hub-1M} dataset in \reftab{table_quality_eval_sarena_1M}.
As expected, the smaller training corpus leads to an overall degradation in performance across metrics, yet our method consistently outperforms prior approaches on all metrics. 
As the dataset size increases, our model improves substantially, whereas other methods show only limited improvement on the text-to-SVG task. 
The results further highlight the effectiveness of our unified multimodal SVG generation model.

\begin{figure}[tbp]
  \centering
  \includegraphics[width=1.0\columnwidth]{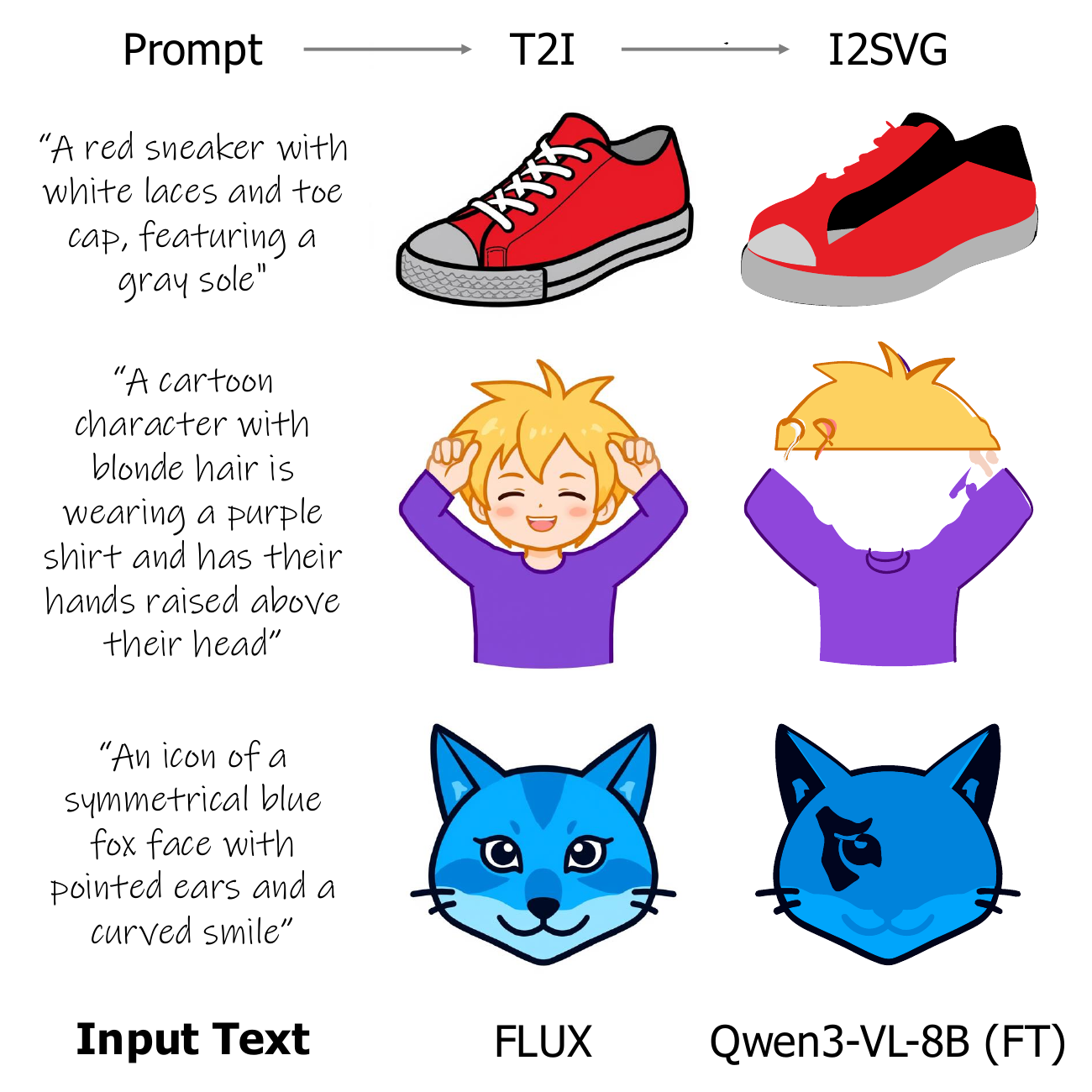}
  \caption{Additional comparisons with Flux + I2SVG pipeline.}
  \label{fig:comp_flux_i2svg}
\end{figure}

\paragraph{Additional Comparisons with Flux + I2SVG.}
For the T2SVG task, we compare against a two-stage baseline that first synthesizes images with FLUX.1-dev \cite{labs2025flux} and then performs image-to-SVG using a VLM-based model, Qwen3-VL-8B \cite{Qwen3-VL}, fine-tuned on \textit{SVG-Hub-5M} dataset. 
We observe cross-model inconsistencies: the synthesized intermediate images differ in style and fine detail from the real SVG images used during I2SVG training, introducing a train-test mismatch for the VLM-based I2SVG model. 
As a result, the I2SVG model does not generalize well to these intermediates, and the resulting SVGs often show reduced alignment with the intermediate images (see \reffig{comp_flux_i2svg}). 
In contrast, our unified multimodal generative model co-generates image and SVG tokens within a single end-to-end architecture, enabling the use of large-scale text-image data and providing visual grounding during SVG decoding.

\begin{table}[tbp]
  \caption{ \textbf{Quantitative comparison with VLM-based baselines on the SArena-Icon benchmark~\cite{wang2025internsvg}. }
  In this setting, \enquote{Ours-7B} and all baselines marked \enquote{FT} are fine-tuned on \textit{SVG-Hub-1M} dataset. 
  }
  \resizebox{\linewidth}{!}{
    \begin{tabular}{l|ccc|cccc}
    \Xhline{1.2pt}

    \multicolumn{1}{c|}{\multirow{2}{*}{Method}} & \multicolumn{3}{c|}{\textbf{Text-to-SVG}}               & \multicolumn{3}{c}{\textbf{Image-to-SVG}}              \\ \cline{2-7} 
    \multicolumn{1}{c|}{}                        & FID $\downarrow$ & FID-C $\downarrow$ & CLIP $\uparrow$ & DINO $\uparrow$ & SSIM $\uparrow$ & LPIPS $\downarrow$ \\ \hline
    StarVector-8B (w/o FT)                       & -                & -                  & -               & 0.871           & 0.623           & 0.206              \\
    StarVector-8B (FT)                           & -                & -                  & -               & 0.895           & 0.742           & 0.182              \\
    LLM4SVG-7B (FT)                              & 20.896           & 7.980              & 22.108          & 0.898           & 0.760           & 0.164              \\
    OmniSVG-3B (w/o FT)                          & 28.292           & 11.318             & 21.679          & 0.894           & 0.756           & 0.186              \\
    OmniSVG-3B (FT)                              & 24.977           & 9.659              & 21.825          & 0.902           & 0.761           & 0.178              \\
    Qwen3-VL-8B (FT)                             & 19.760           & 7.572              & 22.126          & 0.908           & 0.774           & 0.162              \\
    Ours-7B (w/o TTS)                            & {\ul 17.265}     & {\ul 6.388}        & {\ul 22.586}    & {\ul 0.917}     & {\ul 0.808}     & {\ul 0.151}        \\
    Ours-7B (TTS)                                & \textbf{16.952}  & \textbf{6.242}     & \textbf{22.645} & \textbf{0.924}  & \textbf{0.814}  & \textbf{0.145}     \\ 
    
    \Xhline{1.2pt}
    \end{tabular}
  }
  \label{tab:table_quality_eval_sarena_1M}
\end{table}

\begin{figure}[tbp]
  \centering
  \includegraphics[width=1.0\columnwidth]{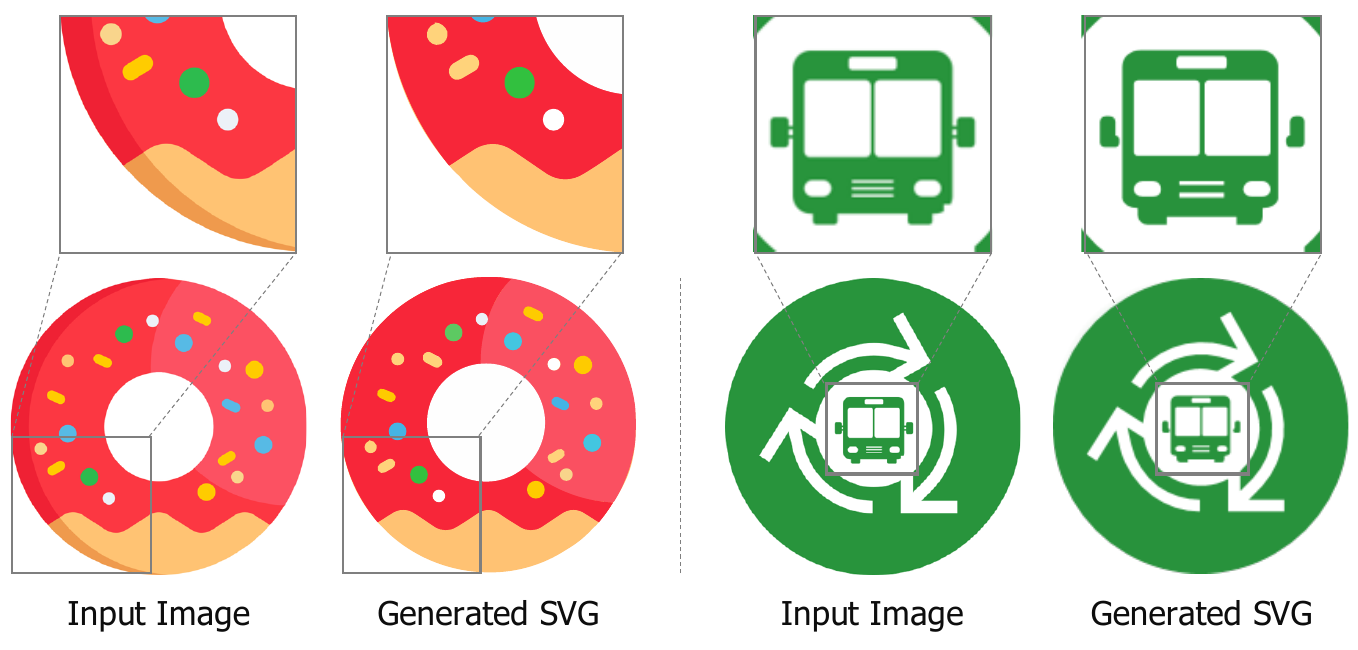}
  \caption{Failure cases.}
  \label{fig:failure_case}
\end{figure}

\section{Limitation}
\label{sec:limitation}

While our model excels at SVG generation, it has limitations. When the input image contains very fine details and rich color variation, the generated SVG may miss small structures and exhibit mild color shifts (see \reffig{failure_case}).
A potential mitigation is a dynamic high-resolution strategy \cite{Qwen2.5-VL} that adaptively increases the number of patches fed to the vision encoder, which can improve the capture of fine details and color consistency. We plan to investigate this in future work.

\section{Details of the captioning pipeline}
\label{sec:captioning_pipeline}

\begin{figure}[H]
  \centering
  \includegraphics[width=1.0\columnwidth]{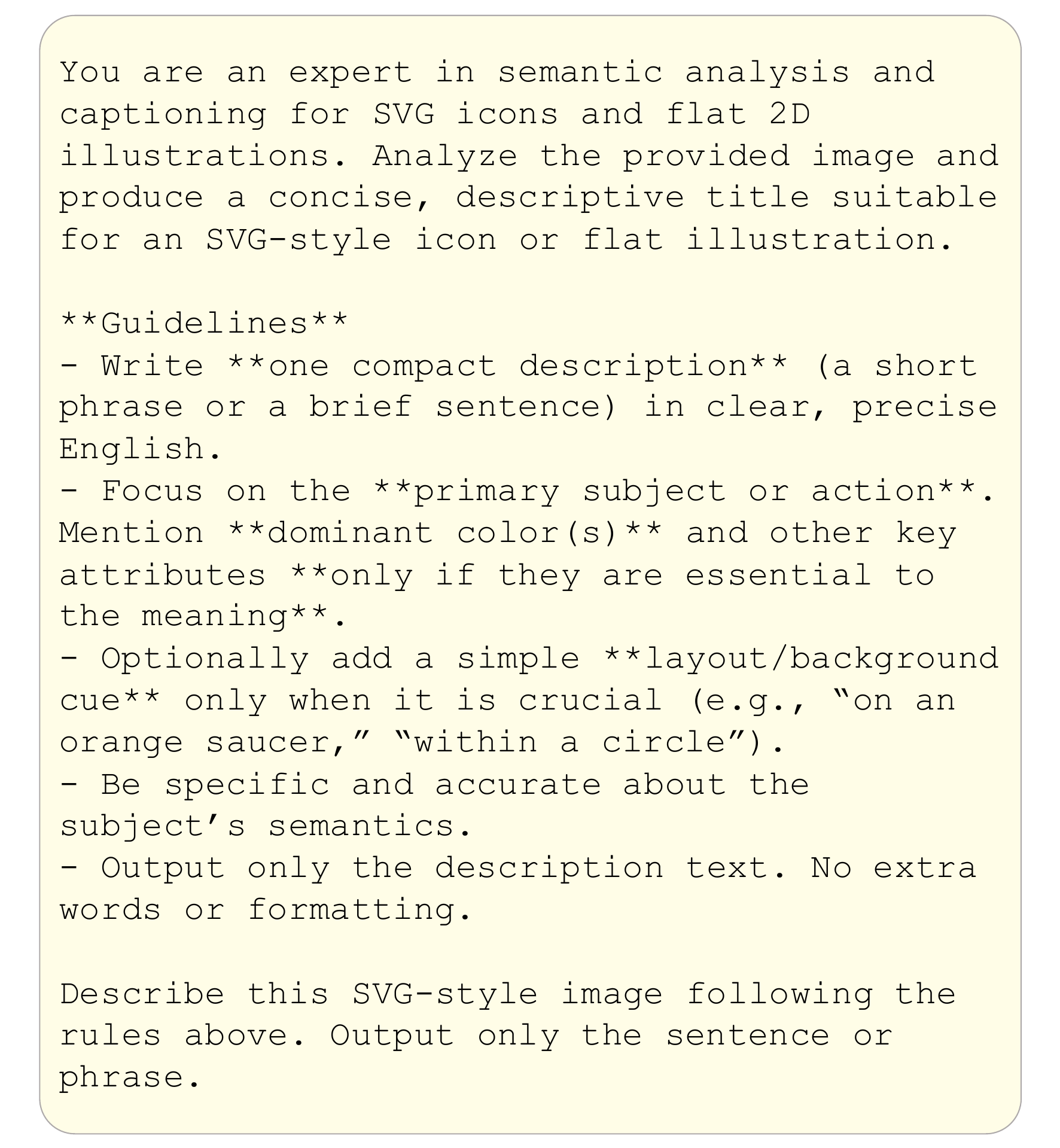}
  \caption{ Short prompt template.}
  \label{fig:caption_template_simple}
\end{figure}

\begin{figure}[H]
  \centering
  \includegraphics[width=1.0\columnwidth]{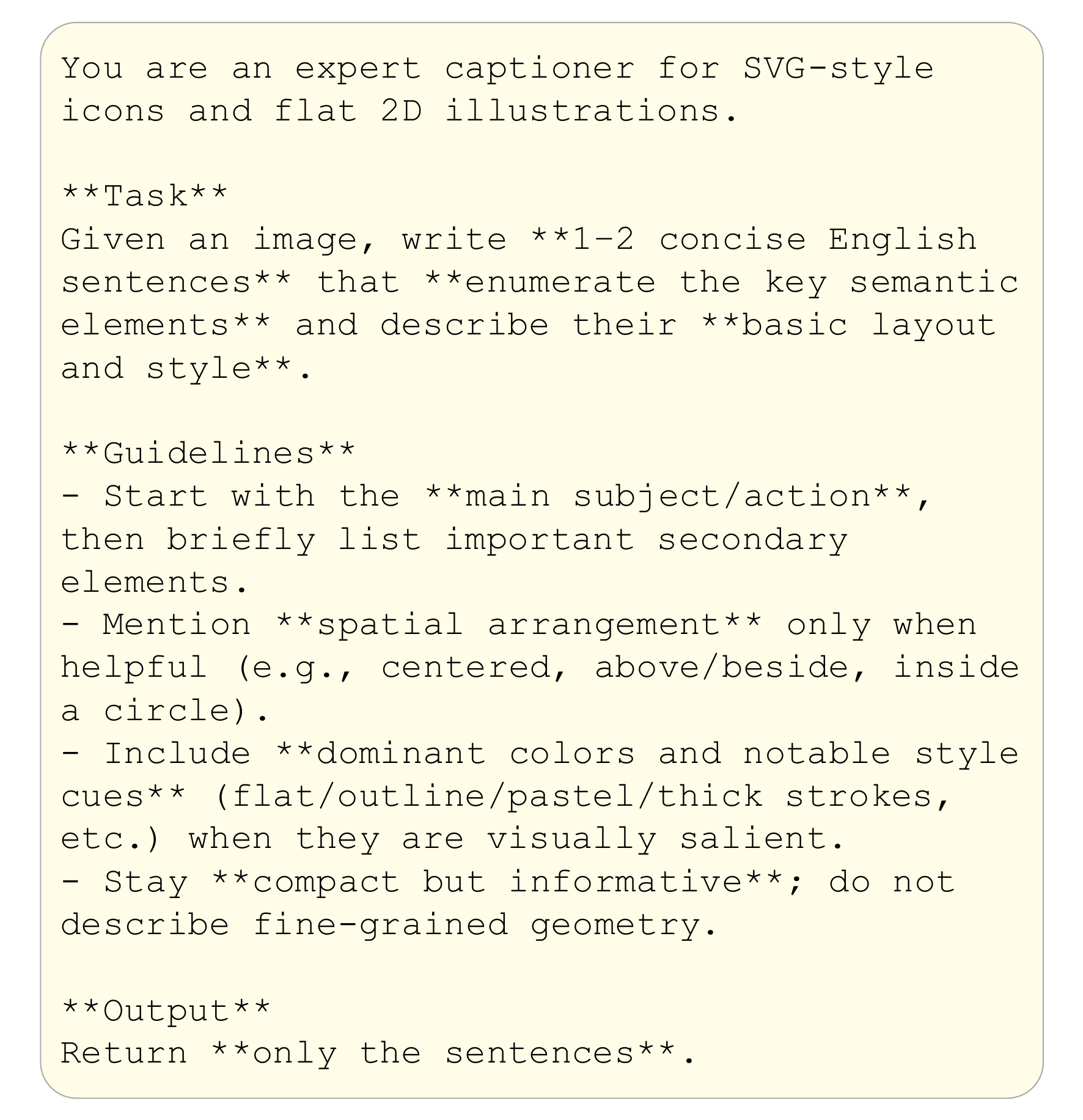}
  \caption{ Medium description template.}
  \label{fig:caption_template_mid}
\end{figure}

\begin{figure}[H]
  \centering
  \includegraphics[width=1.0\columnwidth]{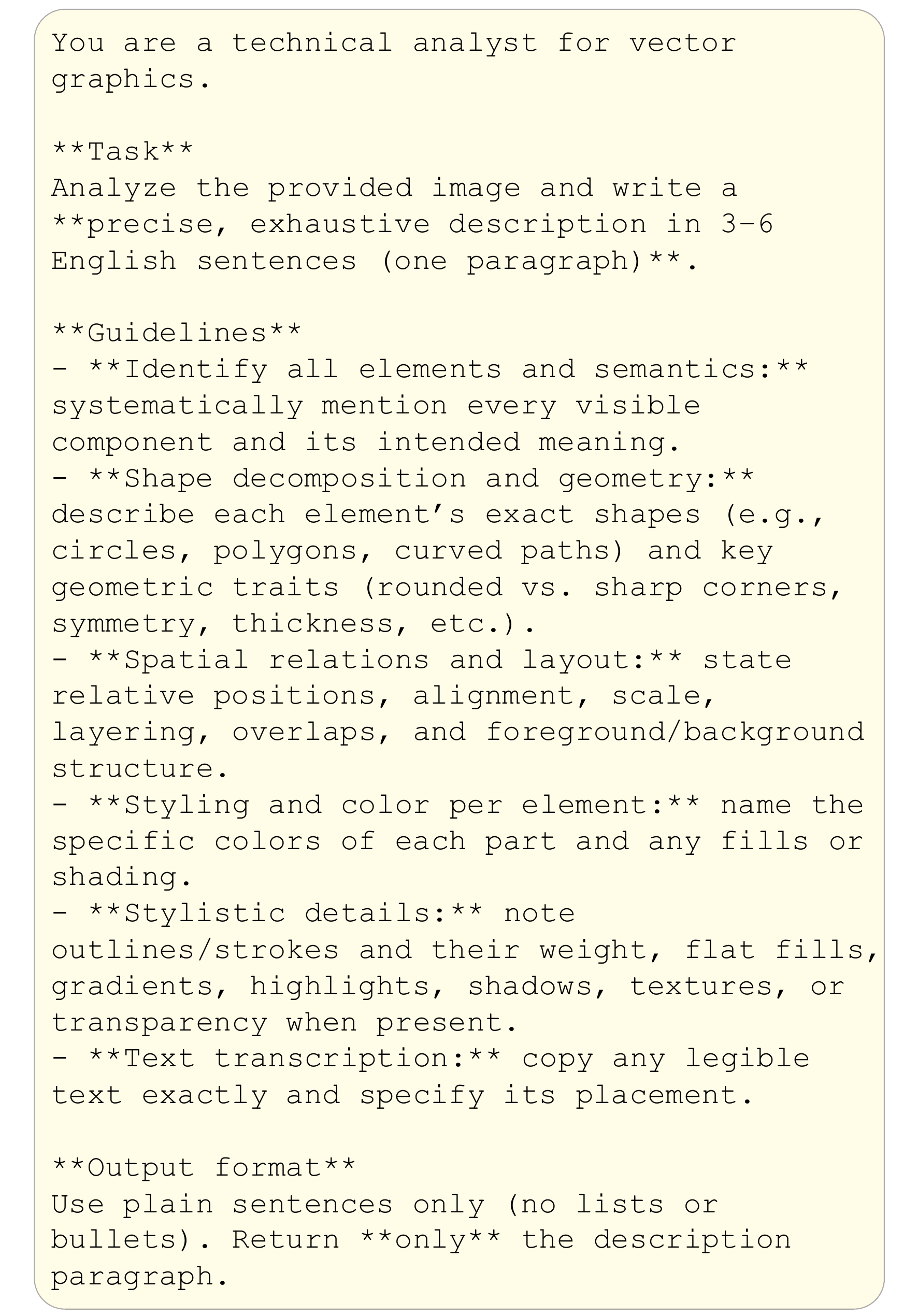}
  \caption{ Detailed annotation template.}
  \label{fig:caption_template_detailed}
\end{figure}

To enable text-to-SVG (T2SVG) training from semantically rich prompts, we rasterize each SVG and use open-source VLMs (InternVL3 \cite{zhu2025internvl3} and Qwen2.5-VL \cite{Qwen2.5-VL}) to generate captions at three levels of detail. We then perform cross-model verification and refinement: a caption produced by one VLM is evaluated against the rendered SVG by the other, which flags inaccuracies and suggests edits; the revised caption is subsequently adopted. The prompt templates for the three levels are provided in \reffig{caption_template_simple}, \reffig{caption_template_mid}, and \reffig{caption_template_detailed}.

\end{sloppypar}
\end{document}